\pdfoutput=1
\documentclass[11pt]{article}

 \usepackage{acl}

\usepackage{times}
\usepackage{latexsym}

\usepackage[T1]{fontenc}
\usepackage[utf8]{inputenc}

\usepackage{microtype}

\usepackage{inconsolata}

\usepackage{multirow}
\usepackage{adjustbox}
\usepackage{booktabs}
\usepackage{arydshln}
\usepackage{todonotes}
\usepackage{color,xcolor,colortbl}
\usepackage[framemethod=TikZ]{mdframed}
\usepackage{pgfplots}
\usepgfplotslibrary{fillbetween}
\usetikzlibrary{patterns}
\usepackage{subcaption}

\pgfplotsset{compat=1.18}

\definecolor{lightgray}{rgb}{0.97,0.97,0.97} % Lighter gray
\definecolor{darkgray}{rgb}{0.8,0.8,0.8} % Lighter gray

\makeatletter
\def\adl@drawiv#1#2#3{%
        \hskip.5\tabcolsep
        \xleaders#3{#2.5\@tempdimb #1{1}#2.5\@tempdimb}%
                #2\z@ plus1fil minus1fil\relax
        \hskip.5\tabcolsep}
\newcommand{\cdashlinelr}[1]{%
  \noalign{\vskip\aboverulesep
           \global\let\@dashdrawstore\adl@draw
           \global\let\adl@draw\adl@drawiv}
  \cdashline{#1}
  \noalign{\global\let\adl@draw\@dashdrawstore
           \vskip\belowrulesep}}
\makeatother

\newcommand{\quotes}[1]{``#1''}

\usepackage{fancyhdr}

\fancypagestyle{firstpage}{
  \fancyhf{} % clear all fields for first page
  \fancyfoot[C]{\thepage} % page number centered
  \fancyfoot[L]{\scriptsize Accepted at WASSA 2024. Official version: \href{https://doi.org/10.18653/v1/2024.wassa-1.6}{10.18653/v1/2024.wassa-1.6}} % left side footer text
}

\title{LLaMA-Based Models for Aspect-Based Sentiment Analysis}

\author{
 \textbf{Jakub \v{S}m\'{i}d\textsuperscript{*}},
 \textbf{Pavel P\v{r}ib\'{a}\v{n}\textsuperscript{*}},
 \textbf{Pavel Kr\'{a}l\textsuperscript{*, $\dagger$}}
\\
\\
University of West Bohemia, Faculty of Applied Sciences, \\
 \textsuperscript{*}Department of Computer Science and Engineering,\\
 \textsuperscript{$\dagger$}NTIS – New Technologies for the Information Society\\
         Univerzitní 2732/8, 301 00 Pilsen, Czech Republic \\
    \{jaksmid, pribanp, pkral\}@kiv.zcu.cz\\
         \url{https://nlp.kiv.zcu.cz}
}

\definecolor{lightblue}{HTML}{5DA5DA}
\definecolor{lightorange}{HTML}{FAA43A}
\definecolor{lightgreen}{HTML}{60BD68}
\definecolor{lightpurple}{HTML}{B276B2}

\definecolor{lightgreenprompt}{HTML}{90EE90}

\mdfsetup{frametitlealignment=\center}

\begin{document}
\maketitle

\thispagestyle{firstpage} % use custom footer on first page

\begin{abstract}
While large language models (LLMs) show promise for various tasks, their performance in compound aspect-based sentiment analysis (ABSA) tasks lags behind fine-tuned models. However, the potential of LLMs fine-tuned for ABSA remains unexplored. This paper examines the capabilities of open-source LLMs fine-tuned for ABSA, focusing on LLaMA-based models. We evaluate the performance across four tasks and eight English datasets, finding that the fine-tuned Orca~2 model surpasses state-of-the-art results in all tasks. However, all models struggle in zero-shot and few-shot scenarios compared to fully fine-tuned ones. Additionally, we conduct error analysis to identify challenges faced by fine-tuned models.
\end{abstract}

\section{Introduction}
Aspect-based sentiment analysis (ABSA) aims to extract detailed sentiment information from text~\citep{absa}. ABSA includes four sentiment elements: aspect term ($a$), aspect category ($c$), opinion term ($o$), and sentiment polarity ($p$). Given the example review \textit{\quotes{The steak was delicious}}, the elements are \textit{\quotes{steak}}, \textit{\quotes{food quality}}, \textit{\quotes{delicious}} and \textit{\quotes{positive}}, respectively.

Initially, ABSA research focused on extracting individual sentiment elements, e.g. aspect term extraction or aspect category detection~\citep{pontiki-etal-2014-semeval}. Recent research has transitioned towards compound tasks involving multiple sentiment elements, such as aspect sentiment triplet extraction (ASTE)~\citep{Peng_Xu_Bing_Huang_Lu_Si_2020}, target aspect category detection (TASD)~\citep{tasd}, aspect category opinion sentiment (ACOS)~\citep{cai-etal-2021-aspect}, and aspect sentiment quad prediction (ASQP)~\citep{zhang-etal-2021-aspect-sentiment}. Table~\ref{tab:absa-tasks} shows the output formats of these ABSA tasks.

\begin{table}[ht!]
    \centering
    \begin{adjustbox}{width=0.95\linewidth}
        \begin{tabular}{lll}
            \toprule
            \textbf{Task} & \textbf{Output} & \textbf{Example output}                                                                                                                 \\ \midrule
            ASTE                   & \{($a$, $o$, $p$)\}       & \{(\quotes{steak}, \quotes{delicious}, POS)\}              \\
            TASD                   & \{($a$, $c$, $p$)\}      & \{(\quotes{steak}, food quality, POS)\}                                        \\
            ACOS                   & \{($a$, $c$, $o$, $p$)\} & \{(\quotes{steak}, food quality, \quotes{delicious}, POS)\}                    \\
            ASQP                   & \{($a$, $c$, $o$, $p$)\} & \{(\quotes{steak}, food quality, \quotes{delicious}, POS)\} \\ \bottomrule
            \end{tabular}
    \end{adjustbox}
    \caption{Output format for selected ABSA tasks for a review: \textit{\quotes{The steak was delicious}}. ACOS focuses on implicit aspect and opinion terms in contrast to ASQP.}
	\label{tab:absa-tasks}
\end{table}

Modern ABSA research often utilizes pre-trained language models, mainly focusing on sequence-to-sequence models. Compound ABSA tasks are typically formulated as text generation problems~\citep{zhang-etal-2021-towards-generative, zhang-etal-2021-aspect-sentiment, gao-etal-2022-lego, hu-etal-2022-improving-aspect, gou-etal-2023-mvp}, which allows to solve compound ABSA tasks simultaneously. 

Lately, large language models (LLMs), such as ChatGPT~\citep{chatgpt-2022}, LLaMA 2~\citep{touvron2023llama} and Orca 2~\citep{mitra2023orca}, have made significant progress across various natural language processing tasks. However, more traditional approaches that fine-tune Transformer-based models with sufficient data have shown superior performance over ChatGPT in compound ABSA tasks~\citep{zhang2023sentiment, gou-etal-2023-mvp}. Additionally, fine-tuning LLMs on a single GPU is challenging due to their large number of parameters. Techniques like QLoRA~\citep{dettmers2023qlora} address this challenge using a quantized 4-bit frozen backbone LLM with a small set of learnable LoRA weights~\citep{hu2021lora}. However, studies have yet to explore the capabilities of fine-tuned open-source LLMs for ABSA.

This paper examines the unexplored potential of LLaMA-based models fine-tuned for English ABSA alongside their performance in zero-shot and few-shot scenarios. Our key contributions include: 1) Introducing the capabilities of fine-tuned LLaMA-based models for ABSA. 2) Conducting a comparative analysis of two LLaMA-based models against state-of-the-art results across four ABSA tasks and eight datasets. 3) Evaluating models' performance in zero-shot, few-shot, and fine-tuning scenarios, demonstrating the superior performance of the fine-tuned Orca 2 model, surpassing state-of-the-art results across all datasets and tasks. 4) Presenting error analysis of the top-performing model.\footnote{Code and datasets are available at \url{https://github.com/biba10/LLaMA-ABSA}.}

\section{Related Work}
Early ABSA studies focused on predicting one or two sentiment elements~\citep{liu-etal-2015-fine, zhou2015representation, he-etal-2019-interactive, cai-etal-2020-aspect} before progressing to more complex tasks involving triplets and quadruplets, such as ASTE~\citep{Peng_Xu_Bing_Huang_Lu_Si_2020}, TASD~\citep{tasd}, ASQP~\citep{zhang-etal-2021-aspect-sentiment} and ACOS~\citep{cai-etal-2021-aspect}.

Recent ABSA research focuses primarily on text generation initiated by GAS~\citep{zhang-etal-2021-towards-generative}. PARAPHRASE~\citep{zhang-etal-2021-aspect-sentiment} converts labels to natural language. LEGO-ABSA~\citep{gao-etal-2022-lego} explores multi-tasking, DLO~\citep{hu-etal-2022-improving-aspect} optimizes element ordering, MVP~\citep{gou-etal-2023-mvp} combines differently ordered outputs, and \citet{scaria2023instructabsa} adopt instruction tuning.

\citet{gou-etal-2023-mvp} and \citet{zhang2023sentiment} show that ChatGPT struggles with compound ABSA tasks in zero-shot and few-shot settings. \citet{simmering2023large} report promising results with close-source LLMs for a single simple ABSA task.

\section{Experimental Setup}
We employ the 7B and 13B versions of LLaMA~2~\citep{touvron2023llama} and Orca~2~\citep{mitra2023orca} models from the HuggingFace Transformers library\footnote{\url{https://github.com/huggingface/transformers}}~\citep{wolf-etal-2020-transformers}. LLaMA~2 offers models of various sizes tailored for dialogue tasks, building upon the LLaMA framework~\citep{touvron2023llama1}. Orca~2 extends this collection with enhanced reasoning capabilities.

\subsection{Experimental Details}
For fine-tuning, we follow recommendations from \citet{dettmers2023qlora} and use QLoRA with the following settings: 4-bit NormalFloat (NF4) with double quantization and bf16 computation datatype, batch size of 16, constant learning rate of 2e-4, AdamW optimizer~\citep{loshchilov2019decoupled}, LoRA adapters~\citep{hu2021lora} on all linear Transformer block layers, and LoRA $r=64$ and $\alpha=16$. We fine-tune the models for up to 5 epochs and choose the best-performing model based on validation loss. Following \citet{mitra2023orca}, we compute loss only on tokens generated by the model, excluding the prompt with instructions.

For zero-shot and few-shot experiments, we use 4-bit quantization of the models. Preliminary experiments indicated that 4-bit quantized models performed similarly to 8-bit quantized models and non-quantized models.

All experiments, including zero-shot and few-shot scenarios, employ greedy search decoding and are conducted on an NVIDIA A40 with 48 GB GPU memory.

% We use QLoRA for fine-tuning and 4-bit quantization for zero- and few-shot experiments. Appendix~\ref{app:experimets} describes the detailed experimental setup.

\subsection{Evaluation Metrics}
We use micro F1-score as the primary evaluation metric, chosen based on related work, and report average results from 5 runs with different seeds. We consider a predicted sentiment tuple correct only if all its elements exactly match the gold tuple.

\subsection{Tasks \& Datasets}
We evaluate the LLMs on four tasks: two involving quadruplets (ASQP and ACOS) and two involving triplets (TASD and ASTE). We select two datasets for each task and use the same data splits as previous works for a fair comparison. Table~\ref{tab:absa-tasks} displays the output targets for each task.

We use Rest15 and Rest16 datasets for ASQP in the restaurant domain, initially introduced in SemEval tasks~\citep{pontiki-etal-2015-semeval, pontiki-etal-2016-semeval}, later aligned and supplemented by \citet{zhang-etal-2021-aspect-sentiment}. For ACOS, we employ ACOS-Rest and ACOS-Lap datasets from \citet{cai-etal-2021-aspect}, focusing on implicit aspects and opinions and providing comprehensive evaluation. We use the dataset from \citet{xu-etal-2020-position} and \citet{tasd} for ASTE and TASD, respectively. Table~\ref{tab:data_stats} shows the detailed data statistics. ASTE datasets are the only ones that do not include implicit sentiment elements.

\begin{table*}[ht!]
    \centering
    \begin{adjustbox}{width=\linewidth}
        \begin{tabular}{@{}llcccccccc@{}}
        \toprule
                               &             & \multicolumn{2}{c}{\textbf{ASQP}}                       & \multicolumn{2}{c}{\textbf{ACOS}}                  & \multicolumn{2}{c}{\textbf{TASD}}                       & \multicolumn{2}{c}{\textbf{ASTE}}                       \\ \cmidrule(lr){3-4} \cmidrule(lr){5-6} \cmidrule(lr){7-8} \cmidrule(lr){9-10}
                               &             & \multicolumn{1}{c}{Rest15} & \multicolumn{1}{c}{Rest16} & \multicolumn{1}{c}{Lap} & \multicolumn{1}{c}{Rest} & \multicolumn{1}{c}{Rest15} & \multicolumn{1}{c}{Rest16} & \multicolumn{1}{c}{Rest15} & \multicolumn{1}{c}{Rest16} \\ \midrule
        \multirow{4}{*}{Train} & Sentences   & 834                        & 1,264                      & 2,934                   & 1,530                    & 1,120                      & 1,708                      & 605                        & 857                        \\
                               & Tuples      & 1,354                      & 1,989                      & 4,172                   & 2,484                    & 1,654                      & 2,507                      & 1,013                      & 1,394                      \\
                               & Categories  & 13                         & 12                         & 114                     & 12                       & 13                         & 12                         & 0                          & 0                          \\
                               & POS/NEG/NEU & 1,005/315/34               & 1,369/558/62               & 2,583/1,362/227         & 1,656/733/95             & 1,198/403/53               & 1,657/749/101              & 783/205/25                 & 1,015/329/50               \\ \cdashlinelr{1-10}
        \multirow{4}{*}{Dev}   & Sentences   & 209                        & 316                        & 326                     & 171                      & 10                         & 29                         & 148                        & 210                        \\
                               & Tuples      & 347                        & 507                        & 440                     & 261                      & 13                         & 44                         & 249                        & 339                        \\
                               & Categories  & 12                         & 13                         & 71                      & 13                       & 6                          & 9                          & 0                          & 0                          \\
                               & POS/NEG/NEU & 252/81/14                  & 341/143/23                 & 279/137/24              & 180/69/12                & 6/7/0                      & 23/20/1                    & 185/53/11                  & 252/76/11                  \\ \cdashlinelr{1-10}
        \multirow{4}{*}{Test}  & Sentences   & 537                        & 544                        & 816                     & 583                      & 582                        & 587                        & 322                        & 326                        \\
                               & Tuples      & 795                        & 799                        & 1,161                   & 916                      & 845                        & 859                        & 485                        & 514                        \\
                               & Categories  & 12                         & 12                         & 81                      & 12                       & 12                         & 12                         & 0                          & 0                          \\
                               & POS/NEG/NEU & 453/305/37                 & 583/176/40                 & 716/380/65              & 667/205/44               & 454/346/45                 & 611/204/44                 & 317/143/25                 & 407/78/29                  \\ \bottomrule 
        \end{tabular}
    \end{adjustbox}
    \caption{Statistics for each dataset. POS, NEG and NEU denote the number of positive, negative and neutral examples, respectively.}
    \label{tab:data_stats}
\end{table*}

\subsection{Prompting Strategy \& Fine-Tuning}
LLMs show varied responses despite similar prompts~\citep{NEURIPS2021_5c049256, lu-etal-2022-fantastically}. 
% Our study aims to standardize prompts across datasets to evaluate LLMs consistently.
Our goal is to design simple, clear, and straightforward prompts to standardize evaluations across datasets and ensure consistent assessment of LLMs.
% Our study aims to provide consistent, simple prompts across datasets and to evaluate LLMs' general performance.

\begin{figure*}[ht!]
    \centering        
    \begin{mdframed}[backgroundcolor=lightgray, roundcorner=10pt, linewidth=1pt,
        frametitle={\footnotesize Prompt for quadruplet tasks}, frametitlerule=true, frametitlebackgroundcolor=darkgray, frametitlerulewidth=0.2pt, font=\scriptsize,
        frametitleaboveskip=2pt,
        frametitlebelowskip=2pt,
        innerleftmargin=3pt,
        innerrightmargin=3pt,
        innertopmargin=2pt,
        innerbottommargin=2pt,
        ]
                \setlength{\parindent}{0pt}
According to the following sentiment elements definition:
\\[-6pt]

- The \quotes{aspect term} refers to a specific feature, attribute, or aspect of a product or service on which a user can express an opinion. Explicit aspect terms appear explicitly as a substring of the given text. The aspect term might be \quotes{null} for the implicit aspect.
\\[-6pt]

- The \quotes{aspect category} refers to the category that aspect belongs to, and the available categories include: \quotes{ambience general}, \quotes{drinks prices}, \quotes{drinks quality}, \quotes{drinks style\_options}, \quotes{food general}, \quotes{food prices}, \quotes{food quality}, \quotes{food style\_options}, \quotes{location general}, \quotes{restaurant general}, \quotes{restaurant miscellaneous}, \quotes{restaurant prices}, \quotes{service general}.
\\[-6pt]

- The \quotes{sentiment polarity} refers to the degree of positivity, negativity or neutrality expressed in the opinion towards a particular aspect or feature of a product or service, and the available polarities include: \quotes{positive}, \quotes{negative} and \quotes{neutral}. \quotes{neutral} means mildly positive or mildly negative. Quadruplets with objective sentiment polarity should be ignored.
\\[-6pt]

- The \quotes{opinion term} refers to the sentiment or attitude expressed by a user towards a particular aspect or feature of a product or service. Explicit opinion terms appear explicitly as a substring of the given text. The opinion term might be \quotes{null} for the implicit opinion.
\\[-6pt]

Please carefully follow the instructions. Ensure that aspect terms are recognized as exact matches in the review or are \quotes{null} for implicit aspects. Ensure that aspect categories are from the available categories. Ensure that sentiment polarities are from the available polarities. Ensure that opinion terms are recognized as exact matches in the review or are \quotes{null} for implicit opinions.
\\[-6pt]

Recognize all sentiment elements with their corresponding aspect terms, aspect categories, sentiment polarity, and opinion terms in the given input text (review). Provide your response in the format of a Python list of tuples: 'Sentiment elements: [(\quotes{aspect term}, \quotes{aspect category}, \quotes{sentiment polarity}, \quotes{opinion term}), ...]'. Note that \quotes{, ...} indicates that there might be more tuples in the list if applicable and must not occur in the answer. Ensure there is no additional text in the response.

\begin{mdframed}[default, backgroundcolor=lightgray, roundcorner=10pt, linewidth=1pt, linecolor=lightgray, font=\scriptsize, tikzsetting={draw=black,dashed,line width=1pt,dash pattern = on 6pt off 3pt}]
Input: \quotes{\quotes{\quotes{We have gone for dinner only a few times but the same great quality and service is given .}}}
\\[-6pt]

Sentiment elements: [(\quotes{service}, \quotes{service general}, \quotes{positive}, \quotes{great}), (\quotes{dinner}, \quotes{food quality}, \quotes{positive}, \quotes{great quality})]
            \end{mdframed}
            Input: \quotes{\quotes{\quotes{It is n't the cheapest sushi but has been worth it every time .}}}
    \\[-6pt]

    \textbf{Output:} \colorbox{lightgreenprompt}{Sentiment elements: [(\quotes{sushi}, \quotes{food prices}, \quotes{neutral}, \quotes{is n't the cheapest}), (\quotes{sushi}, \quotes{food quality}, \quotes{positive}, \quotes{worth})]}
            \end{mdframed}

    \caption{Prompt for quadruplet tasks (ASQP and ACOS) with example input, expected output in a green box, and one demonstration enclosed in a dashed box. The demonstrations are used solely in few-shot scenarios.}
    \label{fig:prompt}
\end{figure*}
Our prompts define sentiment elements and output format. Sentiment element definitions include the permitted label space, e.g. allowed sentiment polarities and aspect categories. The output format describes the expected structure of model responses, allowing us to decode the responses into our desired format. We supplement the prompts with the first ten training examples for a given task for few-shot learning. We use the same prompts for fine-tuning as for zero-shot experiments. Figure~\ref{fig:prompt} illustrates a prompt for quadruplet tasks. Appendix~\ref{app:prompts} presents the prompts for the triplet tasks.

 During the fine-tuning experiments, we train the model to generate the output in the desired format, as shown in Figure \ref{fig:prompt}.

\section{Results}
Table~\ref{tab:res} shows the results of LLaMA-based models. 

\begin{table*}[ht!]
    \centering
    \begin{adjustbox}{width=0.9\linewidth}

        \begin{tabular}{@{}lrrrrrrrrr@{}}
        \toprule
            \multirow{2}[2]{*}{\textbf{Method}}                           & \multicolumn{2}{c}{\textbf{ASQP}}     & \multicolumn{2}{c}{\textbf{ACOS}}     & \multicolumn{2}{c}{\textbf{TASD}}     & \multicolumn{2}{c}{\textbf{ASTE}}     & \multirow{2}[2]{*}{\textbf{AVG}} \\ 
            \cmidrule(lr){2-3} \cmidrule(lr){4-5} \cmidrule(lr){6-7} \cmidrule(lr){8-9}
                                                                        & R15               & R16               & Lap               & Rest              & R15               & R16               & R15               & R16               &              \\ \midrule
            GAS \footnotesize{\citep{zhang-etal-2021-towards-generative}}      & 45.98             & 56.04             & \multicolumn{1}{c}{-}                 & \multicolumn{1}{c}{-}                 & 60.63             & 68.31             & 60.23             & 69.05             &     \multicolumn{1}{c}{-}         \\
            PARAPHRASE \footnotesize{\citep{zhang-etal-2021-aspect-sentiment}} & 46.93             & 57.93             & 43.51             & 61.16             & 63.06             & 71.97             & 62.56             & 71.70             &   59.85           \\
            LEGO-ABSA \footnotesize{\citep{gao-etal-2022-lego}}                & 46.10             & 57.60             & \multicolumn{1}{c}{-}                 & \multicolumn{1}{c}{-}                 & 62.30             & 71.80             & 64.40             & 69.90             &    \multicolumn{1}{c}{-}          \\
            MvP \footnotesize{\citep{gou-etal-2023-mvp}}                       & 51.04             & \underline{60.39}             & \underline{43.92}             & 61.54             & 64.53             & 72.76             & 65.89             & \underline{73.48}             &    61.69          \\ 
            MvP (multi-task) \footnotesize{\citep{gou-etal-2023-mvp}}          & \underline{52.21}             & 58.94             & 43.84             & 60.36             & 64.74             & 70.18             & \underline{69.44}             & 73.10             &  61.60            \\ \cdashlinelr{1-10}
            ChatGPT (zero-shot) \footnotesize{\citep{gou-etal-2023-mvp}}       & 22.87             & \multicolumn{1}{c}{-}                 & \multicolumn{1}{c}{-}                 & 27.11             & \multicolumn{1}{c}{-}                 & 34.08             & \multicolumn{1}{c}{-}                 & \multicolumn{1}{c}{-}                 &    \multicolumn{1}{c}{-}          \\
            ChatGPT (few-shot) \footnotesize{\citep{gou-etal-2023-mvp}}        & 34.27             & \multicolumn{1}{c}{-}                 & \multicolumn{1}{c}{-}                 & 37.71             & \multicolumn{1}{c}{-}                 & 46.50              & \multicolumn{1}{c}{-}                 & \multicolumn{1}{c}{-}                 &   \multicolumn{1}{c}{-}          \\ \midrule
            Orca 2 7B (zero-shot)                                       & 1.19            & 1.66            & 0.87            & 2.52            & 7.77            & 9.80            & 23.04           & 24.58           &  8.93            \\
            Orca 2 7B (few-shot)                                        & 11.34           & 14.21           & 4.50            & 16.00           & 27.32           & 34.13           & 37.70           & 42.18           & 23.42             \\
            Orca 2 7B                                                   & 51.50 & 58.63 & 43.48 & \underline{63.01} & \underline{69.74} & \underline{76.10} & 65.62 & 73.18 &  \underline{62.66}            \\ \cdashlinelr{1-10}
            Orca 2 13B (zero-shot)                                      & 7.83            & 10.23           & 3.20            & 10.98           & 15.62           & 22.84           & 27.74           & 31.64           &   17.46           \\
            Orca 2 13B (few-shot)                                       & 21.13           & 23.47           & 9.10            & 23.80           & 32.00           & 39.08           & 39.50           & 44.16           &    30.16          \\
            Orca 2 13B                                                  & \textbf{52.29} & \textbf{60.82} & \textbf{44.09} & \textbf{65.80} & \textbf{70.49} & \textbf{78.82} & \textbf{69.91} & \textbf{74.23} &   \textbf{64.56}           \\ \cdashlinelr{1-10}
            LLaMA 2 7B (zero-shot)                                      & 0.80            & 1.85            & 0.05            & 2.39            & 2.28            & 7.45            & 3.47            & 5.00            &   3.21           \\
            LLaMA 2 7B (few-shot)                                       & 11.20           & 17.48           & 2.68            & 26.43           & 28.10           & 33.85           & 38.88           & 45.04           &   25.46           \\
            LLaMA 2 7B                                                  & 42.48 & 55.46 & 36.49 & 57.81 & 64.80 & 71.39 & 57.41 & 67.69 &  56.69            \\ \cdashlinelr{1-10}
            LLaMA 2 13B (zero-shot)                                     & 7.54            & 6.86            & 0.72            & 7.79            & 13.65           & 18.04           & 17.43           & 18.66           &   11.34           \\
            LLaMA 2 13B (few-shot)                                      & 12.08           & 19.37           & 2.36            & 23.08           & 35.22           & 38.80           & 31.49           & 38.06           &  25.06            \\
            LLaMA 2 13B                                                 & 47.16 & 52.98 & 38.44 & 60.92 & 67.70 & 74.08 & 61.95 & 69.95 & 59.15    \\ \bottomrule        
        \end{tabular}
    \end{adjustbox}
    \caption{F1 scores on eight datasets of ASQP, ACOS, TASD, and ASTE tasks, along with the average score. The best results are in \textbf{bold}, and the second-best results are \underline{underlined}.}
    \label{tab:res}
\end{table*}

The results demonstrate the remarkable potential of Orca 2, especially in its 13B version, which surpasses previous benchmarks across all four tasks and eight datasets. Notably, the TASD task shows the most significant improvement, with 6\% and 8\% enhancements for the Rest15 and Rest16 datasets, respectively. While improvements for other tasks are relatively smaller, they remain noteworthy. There are marginal enhancements, within 1\%, for the ASQP and ASTE tasks and the ACOS-Lap dataset. However, the ACOS-Rest dataset sees a significant improvement exceeding 4\%, indicating notable progress. The remarkable advancements in the TASD task suggest that predicting opinion terms not included in the TASD task presents the most significant challenge for these models. The larger Orca 2 achieves a substantial improvement of 2.87\% on average. 

The 7B version of Orca 2 performs similarly to the state-of-the-art (SOTA) for most tasks. However, it falls behind by over 2\% in the Rest15 dataset and ASTE task. Nonetheless, it notably exceeds previous SOTA results for the TASD task by 3–5\%, highlighting the challenge of predicting opinion terms absent in the TASD task. Nevertheless, the smaller Orca 2 performs almost 1\% better on average than the previous best results.

Orca 2 significantly outperforms LLaMA 2, with the smaller Orca 2 model even surpassing the larger LLaMA 2 model, underscoring the superior reasoning capabilities of Orca 2. Additionally, it suggests that opting for more advanced but smaller models may be more beneficial than using larger models with less sophistication. The TASD task is the only task LLaMA 2 outperforms previous SOTA results. Compared to previous SOTA results, on average, the larger version is more than 2\% worse, and the smaller version is 5\% worse.

In zero-shot and few-shot scenarios, both evaluated LLaMA-based models exhibit notably inferior performance compared to their fine-tuned counterparts, particularly in quadruplet tasks. ChatGPT, with significantly more parameters, notably outperforms these models across zero-shot and few-shot scenarios. However, ChatGPT notably underperforms compared to fine-tuned models.

\subsection{Error Analysis}

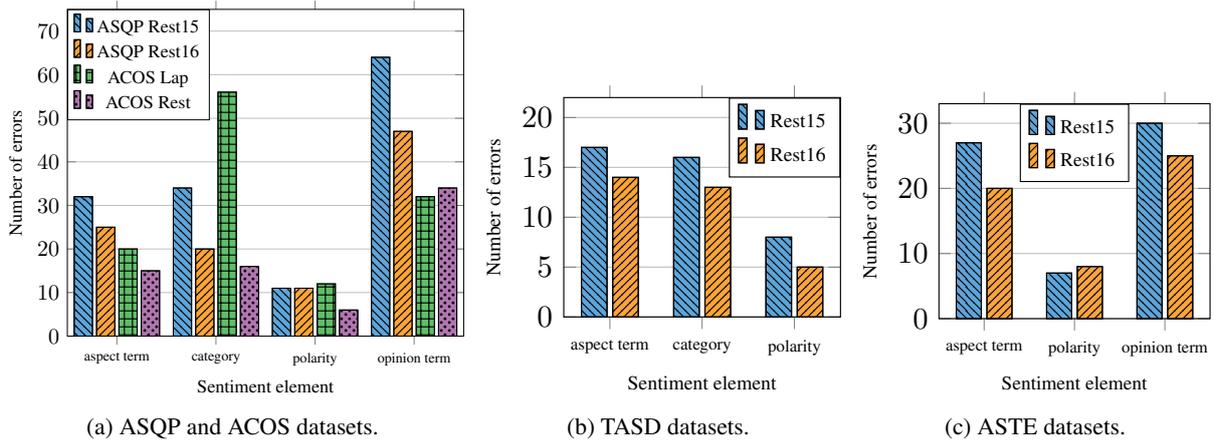
\begin{figure*}[ht!]
    \begin{subfigure}{0.38\textwidth}
        \centering
        \begin{adjustbox}{width=\linewidth}
            \begin{tikzpicture}
                \begin{axis}[
                    ybar,
                    bar width=9pt,
                    xtick={0,1,2,3},
                    ytick={0,10,20,30,40,50,60,70},
                    xticklabels={\scriptsize aspect term,\scriptsize category,\scriptsize polarity,\scriptsize opinion term},
                    ymin=0,
                    ymax=75,
                    xmin=-0.5,
                    xmax=3.5,
                    ymajorgrids=true,
                    ylabel={\footnotesize Number of errors},
                    xlabel={\footnotesize Sentiment element},
                    legend style={at={(0,1)}, anchor=north west, font=\footnotesize},
                    ]
                    % ASQP
                    \addplot[black,fill=lightblue,postaction={pattern=north west lines}] coordinates {(0,32) (1,34) (2,11) (3,64)};
                    \addplot[black,fill=lightorange,postaction={pattern=north east lines}] coordinates {(0,25) (1,20) (2,11) (3,47)};
                     % ACOS
                    \addplot[black,fill=lightgreen,postaction={pattern=grid}] coordinates {(0,20) (1,56) (2,12) (3,32)};
                    \addplot[black,fill=lightpurple,postaction={pattern=crosshatch dots}] coordinates {(0,15) (1,16) (2,6) (3,34)};
                    \legend{ASQP Rest15,ASQP Rest16, ACOS Lap, ACOS Rest}
                \end{axis}
            \end{tikzpicture}
        \end{adjustbox}
       \subcaption{ASQP and ACOS datasets.}
% postaction={pattern=dots}
    \end{subfigure}
    \hfill
    \begin{subfigure}{0.3\textwidth}
        \centering
        \begin{adjustbox}{width=\linewidth}

            \begin{tikzpicture}
                \begin{axis}[
                    ybar,
                    bar width=9pt,
                    width=5cm,
                    xtick={0,1,2},
                    ytick={0,5,10,15,20},
                    xticklabels={\tiny aspect term,\tiny category,\tiny polarity},
                    ymin=0,
                    ymax=22,
                    xmin=-0.5,
                    xmax=2.5,
                    ymajorgrids=true,
                    ylabel={\scriptsize Number of errors},
                    xlabel={\scriptsize Sentiment element},
                    legend style={at={(1,1)}, anchor=north east, font=\scriptsize},
                    ]
                    % TASD
                    \addplot[black,fill=lightblue,postaction={pattern=north west lines}] coordinates {(0,17) (1,16) (2,8)};
                    \addplot[black,fill=lightorange,postaction={pattern=north east lines}] coordinates {(0,14) (1,13) (2,5)};
                    \legend{Rest15, Rest16}
                \end{axis}
            \end{tikzpicture}
        \end{adjustbox}
       \subcaption{TASD datasets.}
    \end{subfigure}
    \hfill
    \begin{subfigure}{0.3\textwidth}
        \centering
        \begin{adjustbox}{width=\linewidth}
            \begin{tikzpicture}
                \begin{axis}[
                    ybar,
                    bar width=9pt,
                    width=5cm,
                    xtick={0,1,2},
                    xticklabels={\tiny aspect term,\tiny polarity,\tiny opinion term},
                    ymin=0,
                    ymax=33,
                    xmin=-0.5,
                    xmax=2.5,
                    ymajorgrids=true,
                    ylabel={\scriptsize Number of errors},
                    xlabel={\scriptsize Sentiment element},
                    legend style={at={(0.5,1)}, anchor=north, font=\scriptsize},
                    ]
                    % ASTE
                    \addplot[black,fill=lightblue,postaction={pattern=north west lines}] coordinates {(0,27) (1,7) (2,30)};
                    \addplot[black,fill=lightorange,postaction={pattern=north east lines}] coordinates {(0,20) (1,8) (2,25)};
                    \legend{Rest15, Rest16}
                \end{axis}
            \end{tikzpicture}
        \end{adjustbox}
        \subcaption{ASTE datasets.}
    \end{subfigure}
    \caption{Number of error types for each dataset.}
    \label{fig:error}
\end{figure*}

To gain insights into the challenges of sentiment prediction, we conduct an error analysis focusing on identifying the most difficult sentiment elements to predict. We manually investigate predictions of 100 random test samples from the best-performing run of Orca~2 with 13B parameters for each dataset. Figure~\ref{fig:error} depicts the results of the error analysis.
% , while Appendix~\ref{app:error} presents comprehensive results for all tasks, with similar conclusions.

In most cases, the most challenging element to predict is the opinion term, often comprising multiple words. The model frequently struggles to predict the text span precisely, for instance, predicting \textit{\quotes{mild}} instead of \textit{\quotes{too mild}}. Following closely in difficulty is typically the aspect term, which encounters similar mistakes as opinion terms, but aspect terms are more often just one word, making such errors less frequent. Sentiment polarity proves to be the easiest to predict. However, an exception arises in the ACOS-Lap dataset, where the aspect category emerges as the most challenging due to the extensive category variety of the dataset (81 categories in the test set, compared to only 12 in the restaurant datasets).

The model also occasionally confuses semantically similar aspect categories, such as \textit{\quotes{restaurant general}} with \textit{\quotes{restaurant miscellaneous}} or \textit{\quotes{keyboard usability}} with \textit{\quotes{keyboard general}}.

The most common error considering sentiment polarity is in predicting the \textit{\quotes{neutral}} class, possibly due to imbalanced label distribution, since the \textit{\quotes{neutral}} class is the least frequent in all datasets.

In contrast to observations made by \citet{zhang-etal-2021-aspect-sentiment}, we did not encounter errors related to text generation, such as generating words for aspect or opinion terms that are absent in the original text.

Additionally, we identified mistakes in the dataset labels. For example, in the ACOS-Rest dataset, the aspect \textit{\quotes{service}} in the sentence \textit{\quotes{worst service i ever had}} is labelled as \textit{\quotes{positive}}, despite being clearly \textit{\quotes{negative}}, a prediction the model also makes correctly. Similarly, we noticed inconsistencies in the datasets, such as in the sentence \textit{\quotes{One of the best hot dogs I have ever eaten}}, where the expression \textit{\quotes{hot dogs}} is not labelled as an aspect term for the \textit{\quotes{food quality}} category; instead, it is labelled as an implicit aspect term (\textit{\quotes{NULL}}), contrary to other examples. These labelling errors could potentially negatively impact the final scores of evaluated models.

\section{Conclusion}
This paper presents a comprehensive evaluation of LLaMA-based models for compound ABSA tasks. We show that these models underperform in zero-shot and few-shot scenarios compared to smaller models fine-tuned specifically for ABSA. However, we demonstrate that fine-tuning the LLaMA-based models for ABSA significantly improves their performance, and the best model outperforms previous state-of-the-art results on all eight datasets and four tasks. Error analysis reveals that predicting opinion terms is generally the most challenging for the evaluated models.

\section*{Acknowledgements}
This work has been partly supported by the OP JAC project DigiTech no. CZ.02.01.01/00/23\_021/0008402 and by the Grant No. SGS-2022-016 Advanced methods of data processing and analysis.
Computational resources were provided by the e-INFRA CZ project (ID:90254), supported by the Ministry of Education, Youth and Sports of the Czech Republic.

\section*{Limitations}
Results highlight LLaMA-based models' ineffectiveness in compound ABSA tasks in zero-shot and few-shot scenarios. Additionally, their performance in non-English languages remains unclear. Future work could also consider other open-source models based on a different architecture.

\section*{Ethics Statement}
We experiment with well-known datasets used in prior scientific studies, ensuring fair and honest analysis while conducting our work ethically and without harming anybody.

\bibliography{anthology,bibliography}

\begin{thebibliography}{30}
\providecommand{\natexlab}[1]{#1}

\bibitem[{Cai et~al.(2020)Cai, Tu, Zhou, Yu, and Xia}]{cai-etal-2020-aspect}
Hongjie Cai, Yaofeng Tu, Xiangsheng Zhou, Jianfei Yu, and Rui Xia. 2020.
\newblock \href {https://doi.org/10.18653/v1/2020.coling-main.72} {Aspect-category based sentiment analysis with hierarchical graph convolutional network}.
\newblock In \emph{Proceedings of the 28th International Conference on Computational Linguistics}, pages 833--843, Barcelona, Spain (Online). International Committee on Computational Linguistics.

\bibitem[{Cai et~al.(2021)Cai, Xia, and Yu}]{cai-etal-2021-aspect}
Hongjie Cai, Rui Xia, and Jianfei Yu. 2021.
\newblock \href {https://doi.org/10.18653/v1/2021.acl-long.29} {Aspect-category-opinion-sentiment quadruple extraction with implicit aspects and opinions}.
\newblock In \emph{Proceedings of the 59th Annual Meeting of the Association for Computational Linguistics and the 11th International Joint Conference on Natural Language Processing (Volume 1: Long Papers)}, pages 340--350, Online. Association for Computational Linguistics.

\bibitem[{Dettmers et~al.(2023)Dettmers, Pagnoni, Holtzman, and Zettlemoyer}]{dettmers2023qlora}
Tim Dettmers, Artidoro Pagnoni, Ari Holtzman, and Luke Zettlemoyer. 2023.
\newblock \href {https://arxiv.org/abs/2305.14314} {Qlora: Efficient finetuning of quantized llms}.
\newblock \emph{Preprint}, arXiv:2305.14314.

\bibitem[{Gao et~al.(2022)Gao, Fang, Liu, Liu, Liu, Liu, Bao, and Yan}]{gao-etal-2022-lego}
Tianhao Gao, Jun Fang, Hanyu Liu, Zhiyuan Liu, Chao Liu, Pengzhang Liu, Yongjun Bao, and Weipeng Yan. 2022.
\newblock \href {https://aclanthology.org/2022.coling-1.610} {{LEGO}-{ABSA}: A prompt-based task assemblable unified generative framework for multi-task aspect-based sentiment analysis}.
\newblock In \emph{Proceedings of the 29th International Conference on Computational Linguistics}, pages 7002--7012, Gyeongju, Republic of Korea. International Committee on Computational Linguistics.

\bibitem[{Gou et~al.(2023)Gou, Guo, and Yang}]{gou-etal-2023-mvp}
Zhibin Gou, Qingyan Guo, and Yujiu Yang. 2023.
\newblock \href {https://doi.org/10.18653/v1/2023.acl-long.240} {{M}v{P}: Multi-view prompting improves aspect sentiment tuple prediction}.
\newblock In \emph{Proceedings of the 61st Annual Meeting of the Association for Computational Linguistics (Volume 1: Long Papers)}, pages 4380--4397, Toronto, Canada. Association for Computational Linguistics.

\bibitem[{He et~al.(2019)He, Lee, Ng, and Dahlmeier}]{he-etal-2019-interactive}
Ruidan He, Wee~Sun Lee, Hwee~Tou Ng, and Daniel Dahlmeier. 2019.
\newblock \href {https://doi.org/10.18653/v1/P19-1048} {An interactive multi-task learning network for end-to-end aspect-based sentiment analysis}.
\newblock In \emph{Proceedings of the 57th Annual Meeting of the Association for Computational Linguistics}, pages 504--515, Florence, Italy. Association for Computational Linguistics.

\bibitem[{Hu et~al.(2021)Hu, Shen, Wallis, Allen-Zhu, Li, Wang, Wang, and Chen}]{hu2021lora}
Edward~J. Hu, Yelong Shen, Phillip Wallis, Zeyuan Allen-Zhu, Yuanzhi Li, Shean Wang, Lu~Wang, and Weizhu Chen. 2021.
\newblock \href {https://arxiv.org/abs/2106.09685} {Lora: Low-rank adaptation of large language models}.
\newblock \emph{Preprint}, arXiv:2106.09685.

\bibitem[{Hu et~al.(2022)Hu, Wu, Gao, Bai, and Zhao}]{hu-etal-2022-improving-aspect}
Mengting Hu, Yike Wu, Hang Gao, Yinhao Bai, and Shiwan Zhao. 2022.
\newblock \href {https://doi.org/10.18653/v1/2022.emnlp-main.538} {Improving aspect sentiment quad prediction via template-order data augmentation}.
\newblock In \emph{Proceedings of the 2022 Conference on Empirical Methods in Natural Language Processing}, pages 7889--7900, Abu Dhabi, United Arab Emirates. Association for Computational Linguistics.

\bibitem[{Liu et~al.(2015)Liu, Joty, and Meng}]{liu-etal-2015-fine}
Pengfei Liu, Shafiq Joty, and Helen Meng. 2015.
\newblock \href {https://doi.org/10.18653/v1/D15-1168} {Fine-grained opinion mining with recurrent neural networks and word embeddings}.
\newblock In \emph{Proceedings of the 2015 Conference on Empirical Methods in Natural Language Processing}, pages 1433--1443, Lisbon, Portugal. Association for Computational Linguistics.

\bibitem[{Loshchilov and Hutter(2019)}]{loshchilov2019decoupled}
Ilya Loshchilov and Frank Hutter. 2019.
\newblock \href {https://arxiv.org/abs/1711.05101} {Decoupled weight decay regularization}.
\newblock \emph{Preprint}, arXiv:1711.05101.

\bibitem[{Lu et~al.(2022)Lu, Bartolo, Moore, Riedel, and Stenetorp}]{lu-etal-2022-fantastically}
Yao Lu, Max Bartolo, Alastair Moore, Sebastian Riedel, and Pontus Stenetorp. 2022.
\newblock \href {https://doi.org/10.18653/v1/2022.acl-long.556} {Fantastically ordered prompts and where to find them: Overcoming few-shot prompt order sensitivity}.
\newblock In \emph{Proceedings of the 60th Annual Meeting of the Association for Computational Linguistics (Volume 1: Long Papers)}, pages 8086--8098, Dublin, Ireland. Association for Computational Linguistics.

\bibitem[{Mitra et~al.(2023)Mitra, Corro, Mahajan, Codas, Simoes, Agarwal, Chen, Razdaibiedina, Jones, Aggarwal, Palangi, Zheng, Rosset, Khanpour, and Awadallah}]{mitra2023orca}
Arindam Mitra, Luciano~Del Corro, Shweti Mahajan, Andres Codas, Clarisse Simoes, Sahaj Agarwal, Xuxi Chen, Anastasia Razdaibiedina, Erik Jones, Kriti Aggarwal, Hamid Palangi, Guoqing Zheng, Corby Rosset, Hamed Khanpour, and Ahmed Awadallah. 2023.
\newblock \href {https://arxiv.org/abs/2311.11045} {Orca 2: Teaching small language models how to reason}.
\newblock \emph{Preprint}, arXiv:2311.11045.

\bibitem[{OpenAI(2022)}]{chatgpt-2022}
OpenAI. 2022.
\newblock \href {https://openai.com/blog/chatgpt} {Openai: Introducing chatgpt}.

\bibitem[{Peng et~al.(2020)Peng, Xu, Bing, Huang, Lu, and Si}]{Peng_Xu_Bing_Huang_Lu_Si_2020}
Haiyun Peng, Lu~Xu, Lidong Bing, Fei Huang, Wei Lu, and Luo Si. 2020.
\newblock \href {https://doi.org/10.1609/aaai.v34i05.6383} {Knowing what, how and why: A near complete solution for aspect-based sentiment analysis}.
\newblock \emph{Proceedings of the AAAI Conference on Artificial Intelligence}, 34(05):8600--8607.

\bibitem[{Perez et~al.(2021)Perez, Kiela, and Cho}]{NEURIPS2021_5c049256}
Ethan Perez, Douwe Kiela, and Kyunghyun Cho. 2021.
\newblock \href {https://proceedings.neurips.cc/paper_files/paper/2021/file/5c04925674920eb58467fb52ce4ef728-Paper.pdf} {True few-shot learning with language models}.
\newblock In \emph{Advances in Neural Information Processing Systems}, volume~34, pages 11054--11070. Curran Associates, Inc.

\bibitem[{Pontiki et~al.(2016)Pontiki, Galanis, Papageorgiou, Androutsopoulos, Manandhar, AL-Smadi, Al-Ayyoub, Zhao, Qin, De~Clercq, Hoste, Apidianaki, Tannier, Loukachevitch, Kotelnikov, Bel, Jim{\'e}nez-Zafra, and Eryi{\u{g}}it}]{pontiki-etal-2016-semeval}
Maria Pontiki, Dimitris Galanis, Haris Papageorgiou, Ion Androutsopoulos, Suresh Manandhar, Mohammad AL-Smadi, Mahmoud Al-Ayyoub, Yanyan Zhao, Bing Qin, Orph{\'e}e De~Clercq, V{\'e}ronique Hoste, Marianna Apidianaki, Xavier Tannier, Natalia Loukachevitch, Evgeniy Kotelnikov, Nuria Bel, Salud~Mar{\'\i}a Jim{\'e}nez-Zafra, and G{\"u}l{\c{s}}en Eryi{\u{g}}it. 2016.
\newblock \href {https://doi.org/10.18653/v1/S16-1002} {{S}em{E}val-2016 task 5: Aspect based sentiment analysis}.
\newblock In \emph{Proceedings of the 10th International Workshop on Semantic Evaluation ({S}em{E}val-2016)}, pages 19--30, San Diego, California. Association for Computational Linguistics.

\bibitem[{Pontiki et~al.(2015)Pontiki, Galanis, Papageorgiou, Manandhar, and Androutsopoulos}]{pontiki-etal-2015-semeval}
Maria Pontiki, Dimitris Galanis, Haris Papageorgiou, Suresh Manandhar, and Ion Androutsopoulos. 2015.
\newblock \href {https://doi.org/10.18653/v1/S15-2082} {{S}em{E}val-2015 task 12: Aspect based sentiment analysis}.
\newblock In \emph{Proceedings of the 9th International Workshop on Semantic Evaluation ({S}em{E}val 2015)}, pages 486--495, Denver, Colorado. Association for Computational Linguistics.

\bibitem[{Pontiki et~al.(2014)Pontiki, Galanis, Pavlopoulos, Papageorgiou, Androutsopoulos, and Manandhar}]{pontiki-etal-2014-semeval}
Maria Pontiki, Dimitris Galanis, John Pavlopoulos, Harris Papageorgiou, Ion Androutsopoulos, and Suresh Manandhar. 2014.
\newblock \href {https://doi.org/10.3115/v1/S14-2004} {{S}em{E}val-2014 task 4: Aspect based sentiment analysis}.
\newblock In \emph{Proceedings of the 8th International Workshop on Semantic Evaluation ({S}em{E}val 2014)}, pages 27--35, Dublin, Ireland. Association for Computational Linguistics.

\bibitem[{Scaria et~al.(2023)Scaria, Gupta, Goyal, Sawant, Mishra, and Baral}]{scaria2023instructabsa}
Kevin Scaria, Himanshu Gupta, Siddharth Goyal, Saurabh~Arjun Sawant, Swaroop Mishra, and Chitta Baral. 2023.
\newblock Instructabsa: Instruction learning for aspect based sentiment analysis.
\newblock \emph{arXiv preprint arXiv:2302.08624}.

\bibitem[{Simmering and Huoviala(2023)}]{simmering2023large}
Paul~F Simmering and Paavo Huoviala. 2023.
\newblock Large language models for aspect-based sentiment analysis.
\newblock \emph{arXiv preprint arXiv:2310.18025}.

\bibitem[{Touvron et~al.(2023{\natexlab{a}})Touvron, Lavril, Izacard, Martinet, Lachaux, Lacroix, Rozière, Goyal, Hambro, Azhar, Rodriguez, Joulin, Grave, and Lample}]{touvron2023llama1}
Hugo Touvron, Thibaut Lavril, Gautier Izacard, Xavier Martinet, Marie-Anne Lachaux, Timothée Lacroix, Baptiste Rozière, Naman Goyal, Eric Hambro, Faisal Azhar, Aurelien Rodriguez, Armand Joulin, Edouard Grave, and Guillaume Lample. 2023{\natexlab{a}}.
\newblock \href {https://arxiv.org/abs/2302.13971} {Llama: Open and efficient foundation language models}.
\newblock \emph{Preprint}, arXiv:2302.13971.

\bibitem[{Touvron et~al.(2023{\natexlab{b}})Touvron, Martin, Stone, Albert, Almahairi, Babaei, Bashlykov, Batra, Bhargava, Bhosale, Bikel, Blecher, Ferrer, Chen, Cucurull, Esiobu, Fernandes, Fu, Fu, Fuller, Gao, Goswami, Goyal, Hartshorn, Hosseini, Hou, Inan, Kardas, Kerkez, Khabsa, Kloumann, Korenev, Koura, Lachaux, Lavril, Lee, Liskovich, Lu, Mao, Martinet, Mihaylov, Mishra, Molybog, Nie, Poulton, Reizenstein, Rungta, Saladi, Schelten, Silva, Smith, Subramanian, Tan, Tang, Taylor, Williams, Kuan, Xu, Yan, Zarov, Zhang, Fan, Kambadur, Narang, Rodriguez, Stojnic, Edunov, and Scialom}]{touvron2023llama}
Hugo Touvron, Louis Martin, Kevin Stone, Peter Albert, Amjad Almahairi, Yasmine Babaei, Nikolay Bashlykov, Soumya Batra, Prajjwal Bhargava, Shruti Bhosale, Dan Bikel, Lukas Blecher, Cristian~Canton Ferrer, Moya Chen, Guillem Cucurull, David Esiobu, Jude Fernandes, Jeremy Fu, Wenyin Fu, Brian Fuller, Cynthia Gao, Vedanuj Goswami, Naman Goyal, Anthony Hartshorn, Saghar Hosseini, Rui Hou, Hakan Inan, Marcin Kardas, Viktor Kerkez, Madian Khabsa, Isabel Kloumann, Artem Korenev, Punit~Singh Koura, Marie-Anne Lachaux, Thibaut Lavril, Jenya Lee, Diana Liskovich, Yinghai Lu, Yuning Mao, Xavier Martinet, Todor Mihaylov, Pushkar Mishra, Igor Molybog, Yixin Nie, Andrew Poulton, Jeremy Reizenstein, Rashi Rungta, Kalyan Saladi, Alan Schelten, Ruan Silva, Eric~Michael Smith, Ranjan Subramanian, Xiaoqing~Ellen Tan, Binh Tang, Ross Taylor, Adina Williams, Jian~Xiang Kuan, Puxin Xu, Zheng Yan, Iliyan Zarov, Yuchen Zhang, Angela Fan, Melanie Kambadur, Sharan Narang, Aurelien Rodriguez, Robert Stojnic, Sergey Edunov, and Thomas
  Scialom. 2023{\natexlab{b}}.
\newblock \href {https://arxiv.org/abs/2307.09288} {Llama 2: Open foundation and fine-tuned chat models}.
\newblock \emph{Preprint}, arXiv:2307.09288.

\bibitem[{Wan et~al.(2020)Wan, Yang, Du, Liu, Qi, and Pan}]{tasd}
Hai Wan, Yufei Yang, Jianfeng Du, Yanan Liu, Kunxun Qi, and Jeff~Z. Pan. 2020.
\newblock \href {https://doi.org/10.1609/aaai.v34i05.6447} {Target-aspect-sentiment joint detection for aspect-based sentiment analysis}.
\newblock \emph{Proceedings of the AAAI Conference on Artificial Intelligence}, 34(05):9122--9129.

\bibitem[{Wolf et~al.(2020)Wolf, Debut, Sanh, Chaumond, Delangue, Moi, Cistac, Rault, Louf, Funtowicz, Davison, Shleifer, von Platen, Ma, Jernite, Plu, Xu, Le~Scao, Gugger, Drame, Lhoest, and Rush}]{wolf-etal-2020-transformers}
Thomas Wolf, Lysandre Debut, Victor Sanh, Julien Chaumond, Clement Delangue, Anthony Moi, Pierric Cistac, Tim Rault, Remi Louf, Morgan Funtowicz, Joe Davison, Sam Shleifer, Patrick von Platen, Clara Ma, Yacine Jernite, Julien Plu, Canwen Xu, Teven Le~Scao, Sylvain Gugger, Mariama Drame, Quentin Lhoest, and Alexander Rush. 2020.
\newblock \href {https://doi.org/10.18653/v1/2020.emnlp-demos.6} {Transformers: State-of-the-art natural language processing}.
\newblock In \emph{Proceedings of the 2020 Conference on Empirical Methods in Natural Language Processing: System Demonstrations}, pages 38--45, Online. Association for Computational Linguistics.

\bibitem[{Xu et~al.(2020)Xu, Li, Lu, and Bing}]{xu-etal-2020-position}
Lu~Xu, Hao Li, Wei Lu, and Lidong Bing. 2020.
\newblock \href {https://doi.org/10.18653/v1/2020.emnlp-main.183} {Position-aware tagging for aspect sentiment triplet extraction}.
\newblock In \emph{Proceedings of the 2020 Conference on Empirical Methods in Natural Language Processing (EMNLP)}, pages 2339--2349, Online. Association for Computational Linguistics.

\bibitem[{Zhang et~al.(2021{\natexlab{a}})Zhang, Deng, Li, Yuan, Bing, and Lam}]{zhang-etal-2021-aspect-sentiment}
Wenxuan Zhang, Yang Deng, Xin Li, Yifei Yuan, Lidong Bing, and Wai Lam. 2021{\natexlab{a}}.
\newblock \href {https://doi.org/10.18653/v1/2021.emnlp-main.726} {Aspect sentiment quad prediction as paraphrase generation}.
\newblock In \emph{Proceedings of the 2021 Conference on Empirical Methods in Natural Language Processing}, pages 9209--9219, Online and Punta Cana, Dominican Republic. Association for Computational Linguistics.

\bibitem[{Zhang et~al.(2023)Zhang, Deng, Liu, Pan, and Bing}]{zhang2023sentiment}
Wenxuan Zhang, Yue Deng, Bing Liu, Sinno~Jialin Pan, and Lidong Bing. 2023.
\newblock Sentiment analysis in the era of large language models: A reality check.
\newblock \emph{arXiv preprint arXiv:2305.15005}.

\bibitem[{Zhang et~al.(2021{\natexlab{b}})Zhang, Li, Deng, Bing, and Lam}]{zhang-etal-2021-towards-generative}
Wenxuan Zhang, Xin Li, Yang Deng, Lidong Bing, and Wai Lam. 2021{\natexlab{b}}.
\newblock \href {https://doi.org/10.18653/v1/2021.acl-short.64} {Towards generative aspect-based sentiment analysis}.
\newblock In \emph{Proceedings of the 59th Annual Meeting of the Association for Computational Linguistics and the 11th International Joint Conference on Natural Language Processing (Volume 2: Short Papers)}, pages 504--510, Online. Association for Computational Linguistics.

\bibitem[{Zhang et~al.(2022)Zhang, Li, Deng, Bing, and Lam}]{absa}
Wenxuan Zhang, Xin Li, Yang Deng, Lidong Bing, and Wai Lam. 2022.
\newblock A survey on aspect-based sentiment analysis: tasks, methods, and challenges.
\newblock \emph{IEEE Transactions on Knowledge and Data Engineering}.

\bibitem[{Zhou et~al.(2015)Zhou, Wan, and Xiao}]{zhou2015representation}
Xinjie Zhou, Xiaojun Wan, and Jianguo Xiao. 2015.
\newblock \href {https://doi.org/10.1609/aaai.v29i1.9194} {Representation learning for aspect category detection in online reviews}.
\newblock \emph{Proceedings of the AAAI Conference on Artificial Intelligence}, 29(1).

\end{thebibliography}

\appendix

\section{Prompts}
\label{app:prompts}
Figure~\ref{fig:prompt_tasd} shows the prompt for the TASD task, while Figure~\ref{fig:prompt_aste} presents the prompts for the ASTE task. The prompts are also available in our code.

\begin{figure*}[ht!]
    \centering        
    \begin{mdframed}[backgroundcolor=lightgray, roundcorner=10pt, linewidth=1pt,
        frametitle={\footnotesize Prompt for the TASD task}, frametitlerule=true, frametitlebackgroundcolor=darkgray, frametitlerulewidth=0.2pt, font=\scriptsize,
        frametitleaboveskip=2pt,
        frametitlebelowskip=2pt,
        innerleftmargin=3pt,
        innerrightmargin=3pt,
        innertopmargin=2pt,
        innerbottommargin=2pt,
        ]
                \setlength{\parindent}{0pt}
According to the following sentiment elements definition:
\\[-6pt]

- The \quotes{aspect term} refers to a specific feature, attribute, or aspect of a product or service on which a user can express an opinion. Explicit aspect terms appear explicitly as a substring of the given text. The aspect term might be \quotes{null} for the implicit aspect.
\\[-6pt]

- The \quotes{aspect category} refers to the category that aspect belongs to, and the available categories include: \quotes{ambience general}, \quotes{drinks prices}, \quotes{drinks quality}, \quotes{drinks style\_options}, \quotes{food general}, \quotes{food prices}, \quotes{food quality}, \quotes{food style\_options}, \quotes{location general}, \quotes{restaurant general}, \quotes{restaurant miscellaneous}, \quotes{restaurant prices}, \quotes{service general}.
\\[-6pt]

- The \quotes{sentiment polarity} refers to the degree of positivity, negativity or neutrality expressed in the opinion towards a particular aspect or feature of a product or service, and the available polarities include: \quotes{positive}, \quotes{negative} and \quotes{neutral}. \quotes{neutral} means mildly positive or mildly negative. Triplets with objective sentiment polarity should be ignored.
\\[-6pt]

Please carefully follow the instructions. Ensure that aspect terms are recognized as exact matches in the review or are \quotes{null} for implicit aspects. Ensure that aspect categories are from the available categories. Ensure that sentiment polarities are from the available polarities.
\\[-6pt]

Recognize all sentiment elements with their corresponding aspect terms, aspect categories, and sentiment polarity in the given input text (review). Provide your response in the format of a Python list of tuples: 'Sentiment elements: [(\quotes{aspect term}, \quotes{aspect category}, \quotes{sentiment polarity}), ...]'. Note that \quotes{, ...} indicates that there might be more tuples in the list if applicable and must not occur in the answer. Ensure there is no additional text in the response.

\end{mdframed}
    \caption{Prompt for the TASD task.}
    \label{fig:prompt_tasd}
\end{figure*}

\begin{figure*}[ht!]
    \centering        
    \begin{mdframed}[backgroundcolor=lightgray, roundcorner=10pt, linewidth=1pt,
        frametitle={\footnotesize Prompt for the ASTE task}, frametitlerule=true, frametitlebackgroundcolor=darkgray, frametitlerulewidth=0.2pt, font=\scriptsize,
        frametitleaboveskip=2pt,
        frametitlebelowskip=2pt,
        innerleftmargin=3pt,
        innerrightmargin=3pt,
        innertopmargin=2pt,
        innerbottommargin=2pt,
        ]
                \setlength{\parindent}{0pt}
According to the following sentiment elements definition:
\\[-6pt]

- The \quotes{aspect term} refers to a specific feature, attribute, or aspect of a product or service on which a user can express an opinion. Explicit aspect terms appear explicitly as a substring of the given text.
\\[-6pt]

- The \quotes{opinion term} refers to the sentiment or attitude expressed by a user towards a particular aspect or feature of a product or service. Explicit opinion terms appear explicitly as a substring of the given text.
\\[-6pt]

- The \quotes{sentiment polarity} refers to the degree of positivity, negativity or neutrality expressed in the opinion towards a particular aspect or feature of a product or service, and the available polarities include: \quotes{positive}, \quotes{negative} and \quotes{neutral}. \quotes{neutral} means mildly positive or mildly negative. Triplets with objective sentiment polarity should be ignored.
\\[-6pt]

Please carefully follow the instructions. Ensure that aspect terms are recognized as exact matches in the review. Ensure that opinion terms are recognized as exact matches in the review. Ensure that sentiment polarities are from the available polarities.
\\[-6pt]

Recognize all sentiment elements with their corresponding aspect terms, opinion terms, and sentiment polarity in the given input text (review). Provide your response in the format of a Python list of tuples: 'Sentiment elements: [(\quotes{aspect term}, \quotes{opinion term}, \quotes{sentiment polarity}), ...]'. Note that \quotes{, ...} indicates that there might be more tuples in the list if applicable and must not occur in the answer. Ensure there is no additional text in the response.
\end{mdframed}
    \caption{Prompt for the ASTE task.}
    \label{fig:prompt_aste}
\end{figure*}

\end{document}